# On the diffusion approximation of nonconvex stochastic gradient descent

Wenqing Hu [*]   Chris Junchi Li [†]   Lei Li [‡]   Jian-Guo Liu [‡]

**Abstract**

We study the Stochastic Gradient Descent (SGD) method in nonconvex optimization problems from the point of view of approximating diffusion processes. We prove rigorously that the diffusion process can approximate the SGD algorithm weakly using the weak form of master equation for probability evolution. In the small step size regime and the presence of omnidirectional noise, our weak approximating diffusion process suggests the following dynamics for the SGD iteration starting from a local minimizer (resp. saddle point): it escapes in a number of iterations exponentially (resp. almost linearly) dependent on the inverse stepsize. The results are obtained using the theory for random perturbations of dynamical systems (theory of large deviations for local minimizers and theory of exiting for unstable stationary points). In addition, we discuss the effects of batch size for the deep neural networks, and we find that small batch size is helpful for SGD algorithms to escape unstable stationary points and sharp minimizers. Our theory indicates that one should increase the batch size at later stage for the SGD to be trapped in flat minimizers for better generalization.

## 1 Introduction

Many nonconvex optimization tasks involve finding desirable stationary point. The Stochastic Gradient Descent (SGD) algorithm and its variants enjoy favorable computational and statistical efficiency and are hence popular in these tasks [4, 7, 5]. A central issue is whether the iteration scheme can escape from unstable stationary points including both saddle points and local maximizers. When the objective function has the *strict saddle property* such that all unstable stationary points have a strictly negative eigenvalue, it is shown by Ge et al. [15] that in both unconstrained and constrained cases, SGD with appropriately chosen stepsizes converges to a local minimizer with high probability after suitably many steps.

Using continuous time processes to approximate stochastic algorithms in machine learning has appeared in several works. In [27, 13], the idea of diffusion approximations for SGDs has appeared. The SGD in [27] is limited to a special class while the approach in [13] is based on the study of

[*]Department of Mathematics and Statistics, Missouri University of Science and Technology (formerly University of Missouri, Rolla); e-mail: huwen@mst.edu

[†]Department of Operations Research and Financial Engineering, Princeton University, Princeton, NJ 08544, USA; e-mail: junchil@princeton.edu

[‡]Department of Mathematics, Duke University, Durham, NC, 27708, USA; email: leili@math.duke.edu, jian-guo.liu@duke.edu

semi–groups. Further, recent stochastic analysis works by the first two authors study the SGD and the stochastic heavy–ball method from their approximating diffusion processes. These diffusion processes is also known as the randomly perturbed gradient flow [19] as well as the randomly perturbed dissipative nonlinear oscillator [20].

In this paper, we consider a broadly general unconstrained stochastic optimization problem and develop an analytic framework of generic diffusion approximations, aiming at better understanding and analyzing the global dynamics of nonconvex SGD. To achieve this, we prove rigorously that the dynamics of nonconvex SGD can be approximated by SDEs using the weak form of master equations. The idea is kind of similar to [13] but the proof here does not use semi–groups explicitly. Our diffusion approximation framework for the nonconvex SGD suggests the vital role of randomness in enabling the algorithm to fast escape from unstable stationary points, as well as efficient convergence to a stationary point. The results are summarized in Theorems 2 and 3 in the main text. Besides, the diffusion approximation also provides insights of the effects of batch size in deep learning, discussed in Section 5.

Ge et al. [15] proved that for discrete–time SGD the iterations for escaping from a saddle point is of order $C \cdot \eta^{-2}$, where $C$ absorbs a power of $d$ and polylogarithmic factor of $\eta^{-1}$, so our rate in Theorem 3 implies a much faster rate than [15] for saddle points escaping. Our result suggests a potentially much sharper convergence rate.

It is worth pointing out that our result is not a proof of the $O(\eta^{-1} \log \eta^{-1})$ escaping rate since the diffusion approximation is proved only on a fixed time interval $[0, T]$. On the other hand, on time interval $[0, \infty)$ such an approximation is only valid in a weaker topology (See Appendix A and compare with [25]). Our hope is that the current work can shed some light in the understanding of the dynamics of discrete stochastic approximation algorithms such as SGD.

The rest of the paper is organized as follows. In Section 2, we give a brief introduction to SGD and related background. In Section 3, we introduce the diffusion process that weakly approximates SGD on any finite interval. This diffusion process is our subject of study in the remaining sections. Moreover, we consider an important example of SGD, namely SGD with mini–batch and its corresponding diffusion process. In Section 4, we consider the limit behavior of the diffusion process. The study here gives us the understanding regarding escaping times from unstable stationary points and local minimizers. The effect of batch size is then discussed in Section 5 with the intuition established in Section 4. We conclude that small batch size is helpful for SGD to escape from sharp local minimizers and unstable stationary points. We also propose to decrease the batch size as the iteration goes on. The classical results for simulated annealing suggests that our intuition makes sense. We present numerical evidence to validate our theory in Section 6. Finally, we conclude our paper in Section **??**.

## 2 Background

In this section we prepare for readers the basic settings of the SGD algorithm. For a generally nonconvex stochastic loss function $f(\mathbf{x}; \zeta)$ that is twice differentiable with respect to $\mathbf{x}$, our goal is to solve the following optimization problem

$$\min_{\mathbf{x}} \quad \mathbb{E}\left[f(\mathbf{x}; \zeta)\right],$$

where $\zeta$ is sampled from some distribution $\mathcal{D}$. For notational convenience, we denote

$$F(\mathbf{x}) := \mathbb{E}\left[f(\mathbf{x}; \zeta)\right]. \tag{2.1}$$

At $k^{th}$ iteration, we evaluate the noisy gradient $\nabla f\left(\mathbf{x}^{(k-1)}; \zeta_k\right)$ and update the iteration according to

$$\mathbf{x}^{(k)} = \mathbf{x}^{(k-1)} - \eta \nabla f(\mathbf{x}^{(k-1)}; \zeta_k). \tag{2.2}$$

Here $\eta > 0$ denotes the *step size* or *learning rate*, and the noisy gradient admits randomness $\zeta_k$ that comes from both noise in stochastic gradient and possibly injected additive noise. We assume that $\zeta_i$ ($i \geq 0$) are i.i.d random variables such that $\zeta_k$ is independent of the sigma algebra $\sigma(\mathbf{x}^0, \mathbf{x}^1, \ldots, \mathbf{x}^{k-1})$. It is straightforward to observe that the iteration $\{\mathbf{x}^{(k)}\}$ generated by (2.2) forms a discrete time, time–homogeneous Markov chain. One of the advantages of stochastic gradient method is that it requires little memory and is hence scalable to the "big data" setting. Throughout this paper, we focus on the case that the stochastic function $f(\mathbf{x}; \zeta)$ is twice–differentiable with respect to $\mathbf{x}$.

Let us introduce the following definition:

**Definition 1** (Stationary Point). *We call $\mathbf{x} \in \mathbb{R}^d$ a stationary point if the gradient $\nabla F(\mathbf{x}) = \mathbf{0}$. By investigating the Hessian matrix at point $\mathbf{x}$, we detail the definition in the following two cases:*

(i) *A stationary point $\mathbf{x}$ is (asymptotically) stable if the Hessian matrix $\nabla^2 F(\mathbf{x})$ is positive definite.*

(ii) *A stationary point $\mathbf{x}$ is unstable if the least eigenvalue of Hessian matrix $\nabla^2 F(\mathbf{x})$ is strictly negative.*

It is straightforward to observe that a stable stationary point is always a local minimizer, and an unstable stationary point can be either a local maximizer (when Hessian has only negative eigenvalues) or a saddle point (when Hessian has both negative and nonnegative eigenvalues). As a side remark, results in [8] implies that nondegenerate stationary points are isolated and can hence be analyzed sequentially.

## 3 Diffusion approximation for SGD

In this section we introduce the concept of the diffusion process that will be analyzed throughout the entire paper for the approximation of SGD.

### 3.1 The diffusion approximation

In this subsection, we justify rigorously that the SGD (2.2) can be approximated by SDEs with global weak accuracy $O(\eta)$ (or $O(\eta^2)$) for $k \in [0, T/\eta]$, where $T$ is a fixed number independent of $\eta$. For this purpose, we introduce the pseudo–time $s$ such that the $k$–th step corresponds to time $s_k = k\eta$.

We introduce the following subset of $C^m$ smooth functions equipped with its natural norm:

$$C_b^m(\mathbb{R}^d) = \left\{ f \in C^m(\mathbb{R}^d) \ \Big| \ \|f\|_{C^m} := \sum_{|\alpha| \leq m} |D^\alpha f|_\infty < \infty \right\}. \tag{3.1}$$

We aim to find diffusion approximations for SGD. Namely to find a diffusion process which solves an SDE and whose trajectory is close to the SGD trajectory in a weak sense. Such an idea appeared in [27, 13]. The SGD algorithms considered in [27] are limited to a special class of objective functions while the approach in [13] relies on semi–groups. We will mostly follow the framework in [13] here but the proof in this paper is slightly different.

Consider that the SDE given as

$$d\boldsymbol{X}(s) = b(\boldsymbol{X}(s))\, ds + \sqrt{\eta}\mathbf{S}(\boldsymbol{X}(s))\, d\boldsymbol{B}(s), \tag{3.2}$$

where $b(\mathbf{x})$ is a vector–valued function and $\mathbf{S}(\boldsymbol{X}(s))$ is a matrix–valued function. We now introduce the concept of weak accuracy:

**Definition 2.** *Fix $T > 0$. We say $\{\boldsymbol{X}(k\eta)\}$ approximates the sequence $\{\mathbf{x}^k\}$ with weak order $p > 0$ if for any $\varphi \in C_b^{2(p+1)}$, there exists $C > 0$ and $\eta_0 > 0$ that are independent of $\eta$ (but may depend on $T$, $\varphi$ and its derivatives) such that*

$$|\mathbb{E}\varphi(\mathbf{x}^k) - \mathbb{E}\varphi(\boldsymbol{X}(k\eta))| \leq C\eta^p \text{ for all } 1 \leq n \leq N \text{ and all } \eta \in (0, \eta_0). \tag{3.3}$$

Let us fix a test function $\varphi \in C_b^6$, and define

$$u^k(\mathbf{x}) = \mathbb{E}_\mathbf{x}(\varphi(\mathbf{x}^k)). \tag{3.4}$$

Then by Markov property, we find the following weak form of master equations

$$u^{k+1}(\mathbf{x}) = \mathbb{E}_\mathbf{x}\left(\mathbb{E}_{\mathbf{x}^1}(\varphi(\mathbf{x}^{k+1})|\mathbf{x}^1)\right) = \mathbb{E}_\mathbf{x}(u^k(\mathbf{x}^1)) = \mathbb{E}\left(u^k(\mathbf{x} - \eta\nabla f(\mathbf{x};\zeta))\right). \tag{3.5}$$

Performing Taylor expansion for $u^k$, we construct SDE approximations of the SGD at different orders. Based on this observation, we prove the following theorem which is a refined version of Theorem 1 in [27]. Our proof uses the semi–group method and is relatively new:

**Theorem 1.** *Assume that there exists $C > 0$ such that for any $\zeta \in \mathcal{D}$,*

$$\|f(\mathbf{x};\zeta)\|_{C^7} \leq C.$$

*If we choose*

$$b(\mathbf{x}) = -\nabla F(\mathbf{x}) - \frac{1}{4}\eta \nabla |\nabla F(\mathbf{x})|^2,$$
$$\mathbf{S}(\mathbf{x}) = \sqrt{\mathbf{\Sigma}(\mathbf{x})}, \tag{3.6}$$

*where*

$$\mathbf{\Sigma}(\mathbf{x}) = \mathrm{var}(\nabla f(\mathbf{x};\zeta)). \tag{3.7}$$

*then the solution to SDE* (3.2) *with* $\mathbf{X}(0) = \mathbf{x}^0$ *approximates the sequence* $\{\mathbf{x}^k\}$ *with weak order* 2.

*Proof.* We fix a test function $\varphi \in C_b^6$. Define $u^k$ as in (3.4) and therefore we have (3.5). Introduce

$$u(\mathbf{x}, s) = \mathbb{E}_{\mathbf{x}} \varphi(\mathbf{X}(s)) \tag{3.8}$$

where $\mathbf{X}(s)$ solves the SDE (3.2). Then, $u(\mathbf{x}, s)$ satisfies the Kolmogrov backward equation

$$\partial_s u = Lu := b \cdot \nabla u + \frac{1}{2}\eta \mathbf{\Sigma} : \nabla^2 u. \tag{3.9}$$

By the assumption on $f(\cdot;\zeta)$, $\|\eta \mathbf{\Sigma}\|_{C^6} \leq C(\eta_0) < \infty$ and $\|b\|_{C^6} \leq C$ for $\eta \leq \eta_0$. It follows that

$$\sup_{0 \leq s \leq T} \|u\|_{C^6} \leq C(T, \eta_0) < \infty, \text{ for all } \eta \leq \eta_0.$$

Further, we have the following semi-group expansion estimate:

$$\left| u(\mathbf{x}, (k+1)\eta) - u(\mathbf{x}, k\eta) - \eta Lu(\mathbf{x}, k\eta) - \frac{1}{2}\eta^2 L^2 u(\mathbf{x}, k\eta) \right| < C(\|u\|_{C^6})\eta^3, \tag{3.10}$$

which implies that

$$\left| u(\mathbf{x}, (k+1)\eta) - u(\mathbf{x}, k\eta) - \eta b \cdot \nabla u - \frac{1}{2}\eta^2(\Sigma + bb^T) : \nabla^2 u - \frac{1}{4}\eta^2 \nabla |b|^2 \cdot \nabla u \right| < C\eta^3. \tag{3.11}$$

Now, by Taylor expansion, we have

$$\mathbb{E}\left( u(\mathbf{x} - \eta \nabla f(\mathbf{x}, \zeta), k\eta) \right) = u(\mathbf{x}, k\eta) - \eta \nabla F(\mathbf{x}) \cdot \nabla u(\mathbf{x}, k\eta)$$
$$+ \frac{1}{2}\eta^2 \mathbb{E}(\nabla f(\mathbf{x}, \zeta) \nabla f(\mathbf{x}, \zeta)^T) : \nabla^2 u(\mathbf{x}, k\eta) + C(\|u\|_{C^3})\eta^3. \tag{3.12}$$

Define

$$E^k = \sup_{\mathbf{x}} |u^k(\mathbf{x}) - u(\mathbf{x}, k\eta)|.$$

With the choice of $b(\cdot)$ and $\mathbf{\Sigma}(\cdot)$ in (3.6), we then find using (3.11) and (3.12) that

$$|u(\mathbf{x}, (k+1)\eta) - \mathbb{E}(u(\mathbf{x} - \eta \nabla f(\mathbf{x}, \zeta), k\eta))| \leq C\eta^3, \ \eta \leq \eta_0. \tag{3.13}$$

Using (3.5) and (3.13), we therefore find

$$E^{k+1} \leq \mathbb{E}(|u^k(\mathbf{x} - \eta \nabla f(\mathbf{x}; \zeta)) - u(\mathbf{x} - \eta \nabla f(\mathbf{x}, \zeta), k\eta)|) + C(T, \eta)\eta^3 \leq E^k + C(T, \eta)\eta^3.$$

The claim then follows.

□

In the $O(\eta^2)$ approximation, $-\frac{1}{4}\eta \nabla |\nabla F(\mathbf{x})|^2$ is a small correction to $b(\mathbf{x})$. Throwing away this term reduces the weak error from $O(\eta^2)$ to $O(\eta)$ as mentioned in [27, 13]. Interestingly, if one goes over the proof, one will find that for $O(\eta)$ weak approximation on $[0, T]$, choosing $b(\mathbf{x}) = -\nabla F(\mathbf{x})$ is enough and the explicit choice of the function $\mathbf{S}(\mathbf{x})$ is not important (despite the fact that we need $\mathbf{S}(\mathbf{x})\mathbf{S}^T(\mathbf{x})$ to be non–degenerate). Indeed, as justified in Appendix A, if we set $\mathbf{\Sigma} = 0$, the magnitude of the error is of order $O(\sqrt{\eta})$. The weak error is of order $O(\eta)$ because the mean of the error is $O(\eta)$. Keeping $\sqrt{\eta \mathbf{\Sigma}}$ indeed captures the essential behavior introduced by the noise. Hence, for simplicity, from here on, we will focus on the following approximation:

$$d\mathbf{X}(s) = -\nabla F(\mathbf{X}(s))\,ds + \sqrt{\eta}\mathbf{S}(\mathbf{X}(s))\,d\mathbf{B}(s), \tag{3.14}$$

where $\mathbf{S}(\mathbf{x}) = [\text{var}(\nabla f(\mathbf{x}; \zeta))]^{1/2}$ is a positive semidefinite matrix–valued function and $\mathbf{B}(s)$ is a standard $d$–dimensional Brownian motion. We call the solution $\mathbf{X}(s)$ to SDE (3.14) the *SGD diffusion process* with respect to (2.2). In the rest of this paper, we concentrate our analysis on the continuous process $\mathbf{X}(s)$ that solves (3.14), which gives us more insight about the original discrete–time SGD dynamics.

### 3.2 An example from deep neural networks

In deep learning [17], our goal is to minimize a generally nonconvex loss function in order to learn the weights of a deep neural network. To be concrete, the goal in training deep neural networks given a training set is to solve the following stochastic optimization problem

$$\min_{\mathbf{x} \in \mathbb{R}^d} \quad F(\mathbf{x}) \equiv \frac{1}{M} \sum_{i=1}^{M} f_i(\mathbf{x}).$$

Here, $M$ is the size of the training set and each component $f_i$ corresponds to the loss function for data point $i \in \{1, \ldots, M\}$. $\mathbf{x}$ is the vector of weights (or parameters for the neural networks) being optimized. When we use mini–batch training and $m = |B_\zeta|$ is the size of minibatch,

$$f(\mathbf{x}; \zeta) = \frac{1}{m} \sum_{i \in B_\zeta} f_i(\mathbf{x}) = \frac{1}{m} \sum_{i=1}^{m} f_{\zeta_i}(\mathbf{x})$$

with $\zeta = (\zeta_1, \ldots, \zeta_m)$. Here, we assume the sampling is uniform and without replacement; i.e. $\zeta_i$ is picked with equal probability from $\{1, \ldots, M\} \setminus \{\zeta_j; 1 \leq j \leq i - 1\}$ for all $1 \leq i \leq m$. The objective function can be further written as the expectation of a stochastic function

$$\frac{1}{M}\sum_{i=1}^{M} f_i(\mathbf{x}) = \mathbb{E}_\zeta \left( \frac{1}{m} \sum_{i \in B_\zeta} f_i(\mathbf{x}) \right).$$

Hence the SGD algorithm (2.2) for minimizing the objective function $F(\mathbf{x})$ iteratively updates the algorithm as

$$\mathbf{x}^{(k)} = \mathbf{x}^{(k-1)} - \eta \cdot \left( \frac{1}{m_k} \sum_{i \in B_{\zeta^{(k)}}} \nabla f_i(\mathbf{x}^{(k-1)}) \right),$$

where $\eta$ is the constant stepsize and $\zeta^{(k)} = (\zeta_1^{(k)}, \ldots, \zeta_{m_k}^{(k)})$. The subindex $k$ for $m$ means that the batch size may depend on time. This is the classical mini–batch version of the SGD. If $m_k = m$ is a constant and $\{\zeta^{(k)} : k \geq 1\}$ are i.i.d, then $\{\mathbf{x}^{(k)}\}$ forms a discrete time-homogeneous Markov chain.

For this classical mini-batch version of SGD, we can accurately quantify its variance:

**Proposition 1.** *Suppose we draw a batch $B_\zeta$ of $m$ data points uniformly from the entire data set $\{1, \ldots, M\}$ without replacement and estimate the gradient with the averaged sample gradients $(1/m)\sum_{i=1}^{M} \nabla f_i(\mathbf{x}) 1_{i \in B_\zeta}$. The estimator is unbiased and the covariance is*

$$\mathbf{\Sigma}(\mathbf{x}) = \text{var}\left( \frac{1}{m} \sum_{i=1}^{M} \nabla f_i(\mathbf{x}) 1_{i \in B_\zeta} \right) = \left( \frac{1}{m} - \frac{1}{M} \right) \mathbf{\Sigma}_0(\mathbf{x}),$$

*where*

$$\mathbf{\Sigma}_0(\mathbf{x}) = \frac{1}{M-1} \sum_{i=1}^{M} (\nabla F(\mathbf{x}) - \nabla f_i(\mathbf{x}))(\nabla F(\mathbf{x}) - \nabla f_i(\mathbf{x}))^\top,$$

*is the sample covariance matrix of random vector $\nabla f_i(\mathbf{x})$.*

*Proof.* We can rewrite the noisy gradient as

$$\nabla f(\mathbf{x}, \zeta) = \frac{1}{m} \sum_{i=1}^{m} \nabla f_{\zeta_i}(\mathbf{x}),$$

where $\zeta = (\zeta_1, \ldots, \zeta_m)$.

The second moment matrix is given by

$$\mathbb{E} \nabla f(\mathbf{x}, \zeta) \nabla f(\mathbf{x}, \zeta)^T = \frac{1}{m^2} \sum_{1 \leq i,j \leq m} \mathbb{E} f_{\zeta_i} f_{\zeta_j}^T$$

If $i = j$, the expectation is

$$\mathbb{E} f_{\zeta_i} f_{\zeta_j}^T = \frac{1}{M} \sum_{p=1}^{M} \nabla f_p \nabla f_p^T,$$

If $i \neq j$, we have

$$\mathbb{E} f_{\zeta_i} f_{\zeta_j}^T = \frac{1}{M} \sum_{k=1}^{M} \mathbb{E}(\nabla f_{\zeta_j} | \zeta_i = k) \nabla_k f = \frac{1}{M(M-1)} \sum_{j \neq k} \nabla f_j \nabla f_k^T$$

Since $\mathbb{E} \nabla f(\mathbf{x}, \zeta) = \nabla F(\mathbf{x}) = \frac{1}{M} \sum_{i=1}^{M} \nabla f_i(\mathbf{x})$, we find that

$$\text{var}(\nabla f(\mathbf{x}, \zeta)) = \mathbb{E} \nabla f(\mathbf{x}, \zeta) \nabla f(\mathbf{x}, \zeta)^T - \nabla F(\mathbf{x}) \nabla F(\mathbf{x})^T$$
$$= \frac{(m-1)}{m} \frac{1}{M(M-1)} \sum_{j \neq k} \nabla f_j \nabla f_k^T + \frac{1}{m} \frac{1}{M} \sum_{p=1}^{M} \nabla f_p \nabla f_p^T - \frac{1}{M^2} \sum_{j,k=1}^{M} \nabla f_j \nabla f_k^T, \quad (3.15)$$

which simplifies to the expression as in the statement. □

Proposition 1 implies that the diffusion matrix $\mathbf{S}(\mathbf{x})$ in the SGD diffusion process is asymptotically equal to $\sqrt{\mathbf{\Sigma}_0(\mathbf{x})/m}$ when $M$ is large. This computation shows that the SDE approximation for this model is given by

$$d\boldsymbol{X}(s) = -\nabla F(\boldsymbol{X}(s)) \, ds + \sqrt{\eta \left( \frac{1}{m} - \frac{1}{M} \right) \boldsymbol{\Sigma}_0} \, d\boldsymbol{B}(s), \qquad (3.16)$$

The batch size therefore affects the magnitude of the diffusion.

## 4 Limiting behavior via stochastic analysis

Throughout this section, we introduce stochastic analysis theory to study the stochastic iterates which are needed to escape from critical points, in the limiting regime of small stepsize $\eta \to 0^+$. We add a $\eta$ on the superscript of the SGD diffusion process $\boldsymbol{X}^\eta(t)$ to emphasize that the process depends on the stepsize $\eta$.

**Notational Conventions** We denote by $\|\mathbf{u}\|$ the Euclidean norm of a vector $\mathbf{u} \in \mathbb{R}^d$. For a real symmetric matrix $\mathbf{H} \in \mathbb{R}^{d \times d}$, let $\lambda_{\min}(\mathbf{H})$ be its smallest eigenvalue. Fixing a connected open set $D \subset \mathbb{R}^d$, a function $g : D \to \mathbb{R}$ is said to be continuously differentiable, denoted by $g \in C^1(D)$, if it has continuous first–order partial derivatives $\partial g / \partial x_i$. Similarly, for any $m \geq 2$, we say $g$ is $m$ times continuously differentiable, denoted by $g \in C^m(D)$, if all the first order partial derivatives are $C^{m-1}(D)$ functions. Let $\nabla F(\mathbf{x})$ and $\nabla^2 F(\mathbf{x})$ be the gradient vector and Hessian matrix at point $\mathbf{x}$ for a function $F \in C^2(D)$. Finally, a matrix valued function $\mathbf{S} : D \to \mathbb{R}^{d \times d}$ is said to be $C^m(D)$ if each entry of $\mathbf{S}$ is a $C^m(D)$ function.

In this subsection we aim at describing the dynamics near local minimizers and saddle points (Theorems 2 and 3). Recall that $D$ is a bounded connected open set with smooth boundary $\partial D$. Let us first introduce the following

**Definition 3.** *We say a matrix valued function $\mathbf{M}(\mathbf{x}) : D \to \mathbb{R}^{d \times d}$ is uniformly positive definite if $\mathbf{M}(\mathbf{x})$ is positive definite at every point $\mathbf{x} \in D$, and $\inf_{\mathbf{x} \in D} \lambda_{\min}(\mathbf{M}(\mathbf{x})) > 0$. In words, the smallest eigenvalue of $\mathbf{M}(\mathbf{x})$ is positive and is bounded away from $0$.*

Let the hitting time $\mathcal{T}^\eta$ of $\partial D$ be

$$\mathcal{T}^\eta = \inf\{s > 0 : \boldsymbol{X}^\eta(s) \in \partial D\}. \tag{4.1}$$

Also, let $\mathbb{E}_{\mathbf{x}}$ denote the conditional expectation operator on $\boldsymbol{X}^\eta(0) = \mathbf{x}$.

## 4.1 Escape from local minimizers

Suppose without loss of generality that $\mathbf{x}^* = \mathbf{0}$ is a non–degenerate local minimizer (otherwise, consider the shifted function $F(\mathbf{x} + \mathbf{x}^*)$). We conclude the following theorem.

**Theorem 2.** *Consider the SDE (3.14). Suppose the matrix–valued function $\mathbf{S}(\mathbf{x}) \in C^1$, $F(\mathbf{x}) \in C^2$ and $\mathbf{S}(\mathbf{x})\mathbf{S}(\mathbf{x})^T$ is uniformly positive definite. Then for any sufficiently small $\delta > 0$ there exists an open ball $B(\mathbf{0}, \delta) \subset U$ such that for any convex open set $D$ inside $B(\mathbf{0}, \delta)$ containing $\mathbf{x} = \mathbf{0}$, there exists a constant $\bar{V}_D \in (0, \infty)$ depending only on $D$ such that the expected hitting time $\mathcal{T}^\eta$ in (4.1) satisfies*

$$\lim_{\eta \to 0+} \eta \log[\mathbb{E}_{\mathbf{x}} \mathcal{T}^\eta] = \bar{V}_D \text{ for all } \mathbf{x} \in D. \tag{4.2}$$

*Further, we have uniform control of the mean exit time: there exist $\delta_1 \in (0, \delta)$, $C_1, C_2 > 0$ and $\eta_0 > 0$ so that whenever $\eta \leq \eta_0$*

$$C_1 \leq \inf_{\mathbf{x} \in B(\mathbf{0}, \delta_1)} \eta \log[\mathbb{E}_{\mathbf{x}} \mathcal{T}^\eta] \leq \sup_{\mathbf{x} \in B(\mathbf{0}, \delta_1)} \eta \log[\mathbb{E}_{\mathbf{x}} \mathcal{T}^\eta] \leq C_2. \tag{4.3}$$

*In particular, we define $\mathcal{N}^\eta = \mathcal{T}^\eta/\eta$ which is the continuous analogue of iteration steps. Then, there exist $C_3, C_4 > 0$ such that the expected steps needed to escape from a local minimizer satisfies*

$$C_3 \leq \inf_{\mathbf{x} \in B(\mathbf{0}, \delta_1)} \eta \log[\mathbb{E}_{\mathbf{x}} \mathcal{N}^\eta] \leq \sup_{\mathbf{x} \in B(\mathbf{0}, \delta_1)} \eta \log[\mathbb{E}_{\mathbf{x}} \mathcal{N}^\eta] \leq C_4. \tag{4.4}$$

**Remark 1.** *Theorem 2 indicates that on average, the system will wander near the local minimizer for asymptotically $\exp(C\eta^{-1})$ number of steps until an escaping event from local minimizer occurs.*

To prove Theorem 2, we show some auxiliary results. Denote

$$u(\mathbf{x}) = \mathbb{E}_{\mathbf{x}} \mathcal{T}^\eta. \tag{4.5}$$

By [10, Corollary 5.7.4], $u(\mathbf{x})$ satisfies the following elliptic PDE with Dirichlet boundary condition

$$\begin{cases} Lu = -1, \ \mathbf{x} \in D, \\ u = 0, \ \mathbf{x} \in \partial D, \end{cases} \tag{4.6}$$

where $L$ is the generator of the diffusion process given by

$$L = \frac{\eta}{2} \sum_{ij} (\mathbf{S}(\mathbf{x})\mathbf{S}(\mathbf{x})^\top)_{ij} \frac{\partial^2}{\partial x_i \partial x_j} - \nabla F(\mathbf{x}) \cdot \nabla. \tag{4.7}$$

The following lemma is useful to us:

**Lemma 1.** *If there exist a function $\psi \in C^2(U) \cap C(\bar{U})$ with $\psi \geq 0$, $\|\psi\|_\infty > 0$, $\psi = 0, \mathbf{x} \in \partial U$ for some open, connected set $U \subset D$ and a positive number $\mu > 0$ such that $-L\psi \leq \mu\psi$, $x \in U$, then*

$$u(\mathbf{x}) \geq \frac{\psi}{\mu\|\psi\|_\infty}, \quad \mathbf{x} \in U.$$

*In particular, suppose $\mu_1$ is the principal eigenvalue of $-L$, then*

$$\|u\|_\infty \geq \frac{1}{\mu_1}.$$

*Proof.* Consider

$$v = u - \frac{\psi}{\mu\|\psi\|_\infty}.$$

Then, $v \geq 0, \mathbf{x} \in \partial U$. Also,

$$-Lv = 1 + \frac{L\psi}{\mu\|\psi\|_\infty} \geq 1 - \frac{\psi}{\|\psi\|_\infty} \geq 0.$$

Then, $v \geq 0$ for $\mathbf{x} \in U$ by maximum principle.

Picking $\mu = \mu_1$ and $\psi$ to be the principal eigenfunction, we obtain the second claim. □

**Lemma 2.** *If $D$ is an open set that contains the non-degenerate local minimizer $\mathbf{x} = \mathbf{0}$ such that there exists $\gamma > 0$ satisfying $\nabla F(\mathbf{x}) \cdot \mathbf{x} > \gamma|\mathbf{x}|^2$ for all $\mathbf{x} \in D$, then*

$$\liminf_{\eta \to 0} \eta \sup_{\mathbf{x} \in D} \log \mathbb{E}_\mathbf{x} \mathcal{T}^\eta > 0. \tag{4.8}$$

*Proof.* By [11, Theorem 4.4],

$$\liminf_{\eta \to 0+} \eta \log(1/\mu_1) > 0,$$

where $\mu_1$ is the principal eigenvalue mentioned in Lemma 1.

Then, by Lemma 1 , we have

$$\liminf_{\eta \to 0} \eta \sup_{\mathbf{x} \in D} \log \mathbb{E}_\mathbf{x} \mathcal{T}^\eta \geq \liminf_{\eta \to 0} \eta \log\left(\frac{1}{\mu_1}\right) > 0. \tag{4.9}$$

The claim follows. □

*Proof of Theorem 2.* This theorem is a natural consequence of the classical Freidlin–Wentzell's large deviation theory. Since $\mathbf{x} = 0$ is a nondegenerate local minimum, we are able to pick $\delta > 0$ such that $\nabla F(\mathbf{x}) \cdot \mathbf{x} > \gamma|\mathbf{x}|^2$ for some $\gamma > 0$ whenever $\mathbf{x} \in B(\mathbf{0}, \delta)$. Now, we fix $D \subset B(\mathbf{0}, \delta)$.

Applying [14, Chapter 4, Theorem 4.1], we conclude that there exists $\bar{V}_D \in [0, \infty)$.

$$\lim_{\eta \to 0} \eta \log \mathbb{E}_{\mathbf{x}} \mathcal{T}^\eta = \bar{V}_D < \infty \text{ for all } \mathbf{x} \in D.$$

Furthermore, given any $\sigma > 0$ that is sufficiently small,

$$\sup_{\mathbf{x} \in D} \mathbb{E}_{\mathbf{x}} \mathcal{T}^\eta \leq T \exp((\bar{V}_D + \sigma)/\eta) \tag{4.10}$$

for some $T > 0$ when $\eta$ is sufficiently small.

We claim that $\bar{V}_D$ is strictly positive. Indeed, using Equation (4.10) and Lemma 2, we have that

$$\bar{V}_D \geq \liminf_{\eta \to 0} \eta \sup_{\mathbf{x} \in D} \log \mathbb{E}_{\mathbf{x}} \mathcal{T}^\eta > 0.$$

The first claim follows.

We move onto inequality (4.3). The existence of $C_2 > 0$ follows directly from Equation (4.10). For the existence of $C_1 > 0$, we choose $\delta_1$ so that it satisfies the requirement on Page 228 of [10]. We apply the first inequality on Page 230 in [10] and get for some $T_0 > 0$ that

$$P_{\mathbf{x}}(\mathcal{T}^\eta \leq e^{(\bar{V}_D - \sigma)/\eta}) \leq 4T_0^{-1} e^{-\sigma/2\eta}, \ \forall \mathbf{x} \in B(0, \delta_1)$$

where we have used the fact that the first term in the inequality on Page 230 in [10] is zero since the starting point $\mathbf{X}_0 = \mathbf{x} \in B(\mathbf{0}, \delta_1)$. As a result,

$$\mathbb{E}_{\mathbf{x}} \mathcal{T}^\eta \geq e^{(\bar{V}_D - \sigma)/\eta}(1 - P_{\mathbf{x}}(\mathcal{T}^\eta \leq e^{(\bar{V}_D - \sigma)/\eta})) > \exp\left(\frac{C_1}{\eta}\right) \tag{4.11}$$

uniformly for $\mathbf{x} \in B(\mathbf{0}, \delta_1)$. The last statement is a corollary of what has been just proved. □

**Remarks** We make several remarks below:

(i) $\bar{V}_D$ is called the *quasi-potential* (see [10, Chap. 5]) and is the cost for forcing the system to be at $\mathbf{z} \in \partial D$ starting from $\mathbf{x}^* = \mathbf{0}$. See [1] for a Poisson clumping heuristics analysis for this process. Introduce the set of functions [10, Chap. 5]

$$V_s(z) = \Big\{ u \in L^2(0, s) : \exists \phi \in C[0, s], \phi(s) = \mathbf{z},$$

$$\forall 0 \leq \tau \leq s, \phi(\tau) = \int_0^\tau (-\nabla F(\phi(\xi))) d\xi + \int_0^\tau \mathbf{S}(\phi(\xi)) u(\xi) d\xi \Big\},$$

where $L^2(0, s)$ means square integrable functions on the interval $[0, s]$. Then, the quasi-potential is given by

$$\bar{V}_D = \inf_{\mathbf{z} \in \partial D} \inf_{s > 0} \inf_{u \in V_s(z)} \frac{1}{2} \int_0^s |u(\tau)|^2 d\tau.$$

The quasi-potential clearly depends on how one choose $\mathbf{S}$ (for example, one may multiply a constant on **S** and redefine $\eta$), but Equation 4.2 is valid for any choice of **S**.

(ii) Let $D$ be a region that contains only one saddle point $z^*$ of $F(\cdot)$. The classical *Eyring-Kramers formula* [16] was first rigorously proved in [6, 28] and concludes as $\varepsilon = \sqrt{\eta} \to 0^+$

$$\mathbb{E}_{x^*}\tau_D = \frac{2\pi}{|\lambda_1(z^*)|} \frac{\sqrt{|\det(\nabla^2 F(z^*))|}}{\sqrt{|\det(\nabla^2 F(x^*))|}} \exp\left(\frac{F(z^*) - F(x^*)}{\varepsilon}\right) \left(1 + \mathcal{O}\left(\varepsilon^{1/2} \log\left(\frac{1}{\varepsilon}\right)\right)\right).$$

This includes a prefactor that depends on both the Hessian of the saddle point $z^*$ and local minimizer $x^*$. However in many applications, we are in the regime $d \cdot \eta \to \infty$. For instance in training deep neural networks, we often have $d = 10^6$ and $\eta = 10^{-4}$. The asymptotics of escaping from a local minimizer in the regime of $\eta \to 0, d \cdot \eta \to \infty$ is an interesting case but lacks mathematical theory.

(iii) Here the escaping time from local minimizer is exponentially dependent on the inverse stepsize. The momentum method such as heavy-ball method [20] or Nesterov's accelerated gradient method [29, 30, 31] is widely adopted in deep learning, and it can help faster escaping from both saddle points and local minimizers. The exact characterization on escaping from local minimizer instead of saddle points is left for future research.

## 4.2 Escape from unstable stationary points

For a generic nondegenerate saddle point (or local maximizer) $\mathbf{x}^*$ we are ready to present the following Theorem 3. To simplify the presentation, we continue to assume without loss of generality that $\mathbf{x}^* = \mathbf{0}$. Also, since $\mathbf{H}$ is real symmetric, it has $d$ real eigenvalues denoted by $\lambda_1 \geq \lambda_2 \geq \ldots \geq \lambda_d$. To be convenient, we introduce $\gamma_i = -\lambda_{d-i}$, and hence $\gamma_1 \geq \gamma_2 \geq \ldots \geq \gamma_d$ are the eigenvalues of $-\mathbf{H}$. Recall that $\mathbf{x} = \mathbf{0}$ is a *nondegenerate stationary* point if $\lambda_i \neq 0$. For a nondegenerate minimizer, we clearly have $\lambda_d > 0$ or $\gamma_1 < 0$, and for a nondegenerate unstable point, $\lambda_d < 0$ or $\gamma_1 > 0$.

**Theorem 3.** *Consider the SDE* (3.14). *Let* $D \subset \mathbb{R}^d$ *be a bounded connected open set with smooth boundary containing the stationary point* $\mathbf{0}$. *Suppose* $\mathbf{S}(\mathbf{x}) : D \to \mathbb{R}^{d \times d}$ *is* $C^3$ *and* $\mathbf{S}(\mathbf{x})\mathbf{S}^T(\mathbf{x})$ *is uniformly positive definite, and* $F : D \to \mathbb{R}$ *is* $C^4$. *If* $\mathbf{x}_0 = \mathbf{0}$ *is a nondegenerate unstable point that satisfies* $\gamma_1 > 0$, *and* $\gamma_i \neq 0$ *for any* $1 \leq i \leq d$, *then conditioned on* $\boldsymbol{X}(0) = \mathbf{0}$, *the expected hitting time* $\mathcal{T}^\eta$ *in* (4.1) *satisfies*

$$\lim_{\eta \to 0} \frac{\mathbb{E}_0 \mathcal{T}^\eta}{\log \eta^{-1}} = 0.5\gamma_1^{-1}. \tag{4.12}$$

*Furthermore for any* $\mathbf{x}_0 \in D$, *conditioned on* $\boldsymbol{X}(0) = \mathbf{x}_0$, *the expected hitting time* $\mathcal{T}^\eta$ *in* (4.1) *satisfies*

$$\lim_{\eta \to 0} \frac{\mathbb{E}_{\mathbf{x}_0} \mathcal{T}^\eta}{\log \eta^{-1}} \leq 0.5\gamma_1^{-1}. \tag{4.13}$$

*In particular, we define $\mathcal{N}^\eta = \mathcal{T}^\eta/\eta$ which is the continuous correspondence of iteration steps. Then, the expected steps needed to escape from a saddle point is asymptotically given by*

$$\mathbb{E}\mathcal{N}^\eta \lesssim \frac{1}{2\gamma_1}\eta^{-1}\log(\eta^{-1}) \ as \ \eta \to 0. \quad (4.14)$$

Theorem 3 follows from the classical dynamical system result in [24] and detailed in §B in the Appendix. In addition, the analysis provided in [24] suggests the following interesting phenomenon: if $\mathbf{x}_0$ is a point such that the ODE (4.15) never hits $\partial D$, then as $\eta \to 0^+$, $\boldsymbol{X}(\mathcal{T}^\eta)$ converges to a measure that concentrates on the intersection between $\partial D$ and the trajectory of ODE initialized at a point $\mathbf{x}_0$ that deviates tiny small from $\mathbf{0}$ at the eigendirection of Hessian corresponding to $\gamma_1$.

**Remark** Note that in (4.13), we have inequality. If the gradient flow ODE system

$$\frac{d\boldsymbol{X}(s)}{ds} = -\nabla F(\boldsymbol{X}(s)), \quad \boldsymbol{X}(0) = \mathbf{x}_0, \quad (4.15)$$

satisfies $\theta(\mathbf{x}_0) = \inf\{t : \boldsymbol{X}(t) \in \partial D\} \in (0, \infty)$, then

$$\lim_{\eta \to 0} \mathbb{E}_{\mathbf{x}_0}\mathcal{T}^\eta = \theta(\mathbf{x}_0).$$

For such $\mathbf{x}_0$ the limit in (4.13) is then given by $\lim_{\eta \to 0}\mathbb{E}_{\mathbf{x}_0}\mathcal{T}^\eta/\log\eta^{-1} = 0$. Hence in the case of nondegenerate local maximizer, (4.13) gives the limit 0 for all points but $\mathbf{0}$. However in the case of saddle points, the limit is nonzero for all points on the so-called *stable manifold*.

## 5  Effects of batch size in deep neural networks

In this section, we use the SDE to discuss how batch size affects the behavior of the SGD algorithm for deep neural networks as introduced in §4. In recent years, deep neural network has achieved state-of-the-art performance in a variety of applications including computer vision, natural language processing and reinforcement learning. Training a deep neural network frequently involves solving a nonconvex stochastic optimization problem using SGD algorithms and their variants, which has also raised many interesting theoretical questions. We conclude from Proposition 1 that the variance of randomness in each iterate scales linearly with respect to the inverse batch size, and therefore the randomness level of a small-batch method is higher than its large-batch counterpart. Based on this observation, using the diffusion framework, we are able to explain the effects of batch size in deep learning as follows:

(i) Smaller batch size leads to larger omnidirectional noise, which enables rapid escaping from nondegenerate saddle points, and hence consistently produce good local minimizer. This explains the saddle-point escaping phenomenon exhibited by Dauphin et al. and Keskar et al. in [8, 23].

(ii) In the small but finite $\eta$ regime, SGD with small batch sizes escape sharp minimizers more

easily compared with flat minimizers, so small batch size at the early stage can help SGD to escape sharp minimizers, explaining the phenomenon of escaping from local minimizers observed by Keskar et al. in [23].

(iii) For the SGD to settle down at flat minimizers, which have better generalization property, we need to use larger batch size at later stage. This agrees with the classical results from simulated annealing as discussed in detail below.

Let us now focus on how changing the batch size could possibly lead to better performance using the diffusion approximation. For the convenience of discussion, let us denote

$$\beta(s) = \eta \left( \frac{1}{m(s)} - \frac{1}{M} \right) \sim \frac{\eta}{m(s)}, \qquad \mathbf{S}_0 = \sqrt{\Sigma_0},$$

and the SDE reads

$$d\boldsymbol{X}(s) = -\nabla F(\boldsymbol{X}(s))\,ds + \sqrt{\beta(s)}\,\mathbf{S}_0\,d\boldsymbol{B}.$$

Our motivation is to find a good $\beta(s)$ such that the process converges to the global minimizer fast enough. A possible framework is to solve this problem from a stochastic control viewpoint [12]. However, the resulted Hamilton-Jacobi-Bellman equation is hard to analyze. Here we state one classical result from simulated annealing, and conclude that for convergence to a global minimizer, varying batch size may provide a useful strategy.

**Proposition 2.** *[21] Suppose there exist $R > 0$ and $a > 0$ such that for all $|\mathbf{x}| > R$, $(\mathbf{x}/\|\mathbf{x}\|) \cdot \nabla F(\mathbf{x}) \geq a$, and that there are finitely many local minimizers inside $|\mathbf{x}| \leq R$. Denote $A$ the set of global minimizers of $F(\mathbf{x})$ and $\boldsymbol{X}(s)$ solve the following SDE*

$$d\boldsymbol{X}(s) = -\nabla F(\boldsymbol{X}(s))\,ds + \sqrt{\frac{\gamma}{\log(2+s)}}\,d\boldsymbol{B}(s).$$

*Then there exists $\gamma_0 > 0$ such that for all $\epsilon > 0$ and $\gamma > \gamma_0$,*

$$\lim_{s \to \infty} \mathbb{P}(\boldsymbol{X}(s) \in A^\epsilon) = 1,$$

*where $A^\epsilon$ denotes the $\epsilon$-neighborhood of $A$.*

Proof of Proposition 2 can be found in [21, Theorem 3.3]. This proposition tells us that at the early stage, we should use large $\eta$ and small batch size $m$ to increase the diffusivity so that the process will not be trapped in sharp local minimizers and escape from saddle points faster. At later stage, we may choose large batch size so that the process will cool down into the global minimizers or flat local minimizers. The rule is to set

$$m(s) \approx \min(C \log(s+2)/\eta, m^*),$$

where $m^*$ is the largest batch size one may want to use in optimization. This agrees with the intuition from the previous discussion.

# 6 Numerical experiment

In this section, we set up experiments and present numerical evidences to validate our theory mentioned above. We also propose a conjecture saying that we could manually add noise to the gradient of nonconvex stochastic loss function and increase the randomness level of certain batch size method. Concretely, we add noise to the updated weights and apply the following algorithm to optimize weights of neural network with large batch method. We have

$$\mathbf{x}^{(k)} = \mathbf{x}^{(k-1)} - \eta \cdot \left( \frac{1}{m_k} \sum_{i \in B_k} \nabla f_i(\mathbf{x}^{(k-1)}) + \epsilon \right)$$

where $\epsilon \sim \mathcal{N}(0, \sigma^2 I)$ is a random vector independent of $\mathbf{x}$ and $\sigma$ is a positive constant. We consider a popular and representative configuration used by Keskar et al. [23] which is a fully–connected model for training MNIST data set. The model uses 784 neurons as input layer and use 10 neurons with softmax activate function as output layer. The hidden layers consist of 5 fully connected layers with ReLU activation function followed by batch–normalized layers [22]. For all experiments, we used a batch of 2048 samples as large batch method and 128 samples as small batch method. We used SGD optimizer without momentum and for the conjecture we used modified SGD optimizer. To observe the diffusion process of weights, we calculate Euclidean distances of weight vectors from the initial weight vector for each iteration. We also used the best results from large batch method and small batch method as initial weights to train model for validating the randomness level around local minimum.

As illustrated in Figure 3 and 4, our experiments got similar generalization performance gap with Keskar et al. in [23] which leads to high training accuracy but low test accuracy respectively, particularly by using large batch method. The gap between training accuracy and test accuracy becomes wider as the batch size becomes larger. In Figure 1 and 2, the small batch method's weights go further than large batch method's weights at the same iteration of updated weights. This result is similar with Hoffer et al. in [18]. Different batch size methods have different diffusion rates and the distances between initial weights and weights at the end of training process are different. According to our theory, this phenomenon illustrates that small batch size method has high level of omnidirectional noise which enables weights to random walk with large range of surface of nonconvex loss function. Essentially, the Ghost Batch Normalization [18] which is to acquire statistics on partial batch rather than the full batch statistic increases the randomness level. Therefore, we infer that at the end of training process large batch method tends to converge to sharper minimizers which has lower test accuracy [23] because of low diffusivity. As shown in Figure 5 and 6, we can confirm small batch method has larger diffusion rate than large batch method and by vibrating test accuracy one could get a better result. As illustrated in Figure 5, when the weights come to a flat area relatively, large batch method tends to find the global minimizers or flat local minimizers rather than small batch method with random walk. The only way to find a better result than large batch method by using small batch method is training longer. It increases the probability of getting a good result during a training process. On the contrary, in Figure 6 when the weights come to a sharp area relatively, small batch method make weights escape from sharp minimizers.

As mentioned above, we try to propose a method which could increase the randomness level of certain batch size. The reason why we propose such conjecture is the different diffusivity of large batch method and small batch method and that large batch method is good for parallelizing SGD for Deep learning [9]. It is a good way to speed up training process and decrease the computational gap. We tune the value $\sigma = 0.0054$ to make large batch method's curve similar to small batch method. However, the results are similar using large batch method and have a bit improvement by using SGD with momentum. These phenomena mean that we need more elaborate noise rather than plain spherical noise. At the early stage, we can increase the randomness of large batch method so that weights do not trapped in sharp minimizers and escape from saddle points faster. When weights come to a flat minimizer, we just use normal large batch method to make weight trapped in flat minimizers.

## Acknowledgements

The authors thank Yuhao Tang for assistance of figures and simulation plots. J.-G. Liu and L. Li are supported in part by National Science Foundation (NSF) under award DMS-1514826. C. J. Li is partially supported by RNMS11-07444(KI-Net) during his visit at Duke University.

## References

[1] David Aldous. *Probability Approximations via the Poisson Clumping Heuristic*, volume 77. Springer, 1989.

[2] A. Benveniste, M. Métivier, and P. Priouret. *Adaptive Algorithms and Stochastic Approximations*. Springer, 2012.

[3] Vivek S Borkar. *Stochastic Approximation: A Dynamical Systems Viewpoint*. Cambridge University Press, 2008.

[4] Léon Bottou. Large-scale machine learning with stochastic gradient descent. In *Proceedings of COMPSTAT'2010*, pages 177–186. Springer, 2010.

[5] Léon Bottou, Frank E Curtis, and Jorge Nocedal. Optimization methods for large-scale machine learning. *arXiv preprint arXiv:1606.04838*, 2016.

[6] Anton Bovier, Michael Eckhoff, Véronique Gayrard, and Markus Klein. Metastability in reversible diffusion processes i: Sharp asymptotics for capacities and exit times. *Journal of the European Mathematical Society*, 6(4):399–424, 2004.

[7] Sébastien Bubeck et al. Convex optimization: Algorithms and complexity. *Foundations and Trends® in Machine Learning*, 8(3-4):231–357, 2015.

[8] Yann N Dauphin, Razvan Pascanu, Caglar Gulcehre, Kyunghyun Cho, Surya Ganguli, and Yoshua Bengio. Identifying and attacking the saddle point problem in high-dimensional non-convex optimization. In *Advances in Neural Information Processing Systems*, pages 2933–2941, 2014.

[9] Jeffrey Dean, Greg Corrado, Rajat Monga, Kai Chen, Matthieu Devin, Mark Mao, Marc aurelio Ranzato, Andrew Senior, Paul Tucker, Ke Yang, Quoc V. Le, and Andrew Y. Ng. Large scale distributed deep networks. In F. Pereira, C. J. C. Burges, L. Bottou, and K. Q. Weinberger, editors, *Advances in Neural Information Processing Systems 25*, pages 1223–1231. Curran Associates, Inc., 2012.

[10] Amir Dembo and Ofer Zeitouni. *Large deviations techniques and applications.* Springer-Verlag, Berlin, 2010.

[11] Allen Devinatz, Richard Ellis, and Avner Friedman. The asymptotic behavior of the first real eigenvalue of second order elliptic operators with a small parameter in the highest derivatives. ii. *Indiana University Mathematics Journal*, 23:991–1011, 1974.

[12] Lawrence C Evans. An introduction to mathematical optimal control theory. *Lecture Notes, University of California, Department of Mathematics, Berkeley*, 2005.

[13] Yuanyuan Feng, Lei Li, and Jian-Guo Liu. A note on semi-groups of stochastic gradient descent and online principal component analysis. *Communications in Mathematical Sciences*, To appear.

[14] Mark I Freidlin, Joseph Szücs, and Alexander D Wentzell. *Random Perturbations of Dynamical Systems, Third Edition.* Springer, 2012.

[15] Rong Ge, Furong Huang, Chi Jin, and Yang Yuan. Escaping from saddle points – online stochastic gradient for tensor decomposition. In *Proceedings of The 28th Conference on Learning Theory*, pages 797–842, 2015.

[16] Samuel Glasstone, Keith J. Laidler, and Henry Eyring. *The Theory of Rate Processes.* McGraw-Hill, 1941.

[17] Ian Goodfellow, Yoshua Bengio, and Aaron Courville. *Deep Learning.* MIT Press, 2016. http://www.deeplearningbook.org.

[18] Elad Hoffer, Itay Hubara, and Daniel Soudry. Train longer, generalize better: closing the generalization gap in large batch training of neural networks. In I. Guyon, U. V. Luxburg, S. Bengio, H. Wallach, R. Fergus, S. Vishwanathan, and R. Garnett, editors, *Advances in Neural Information Processing Systems 30*, pages 1729–1739. Curran Associates, Inc., 2017.

[19] Wenqing Hu and Chris Junchi Li. On the fast convergence of random perturbations of the gradient flow. *arXiv preprint arXiv:1706.00837*, 2017.

[20] Wenqing Hu, Chris Junchi Li, and Weijie Su. On the global convergence of a randomly perturbed dissipative nonlinear oscillator. *arXiv preprint arXiv:1712.05733*, 2017.

[21] C.-R. Hwang and S.-J. Sheu. Large-time behavior of perturbed diffusion markov processes with applications to the second eigenvalue problem for Fokker-Planck operators and simulated annealing. *Acta Applicandae Mathematicae*, 19(3):253–295, 1990.

[22] Sergey Ioffe and Christian Szegedy. Batch normalization: Accelerating deep network training by reducing internal covariate shift. In Francis Bach and David Blei, editors, *Proceedings of the 32nd International Conference on Machine Learning*, volume 37 of *Proceedings of Machine Learning Research*, pages 448–456, Lille, France, 07–09 Jul 2015. PMLR.

[23] Nitish Shirish Keskar, Dheevatsa Mudigere, Jorge Nocedal, Mikhail Smelyanskiy, and Ping Tak Peter Tang. On large-batch training for deep learning: Generalization gap and sharp minima. In *ICLR*, 2017.

[24] Yuri Kifer. The exit problem for small random perturbations of dynamical systems with a hyperbolic fixed point. *Israel Journal of Mathematics*, 40(1):74–96, 1981.

[25] H.J. Kushner. A cautionary note on the use of singular perturbation methods for "small noise" modelst. *Stochastics: formerly Stochastics and Stochastics Reports*, 6(2):117–120, 1982.

[26] H.J. Kushner and G. Yin. *Stochastic Approximation and Recursive Algorithms and Applications*. Springer, 2003.

[27] Qianxiao Li, Cheng Tai, and E Weinan. Stochastic modified equations and adaptive stochastic gradient algorithms. In *International Conference on Machine Learning*, pages 2101–2110, 2017.

[28] Robert S Maier and Daniel L Stein. Limiting exit location distributions in the stochastic exit problem. *SIAM Journal on Applied Mathematics*, 57(3):752–790, 1997.

[29] Yurii Nesterov. *Introductory Lectures on Convex Optimization: A Basic Course*, volume 87. Springer, 2013.

[30] Weijie Su, Stephen Boyd, and Emmanuel J Candes. A differential equation for modeling Nesterov's accelerated gradient method: theory and insights. *Journal of Machine Learning Research*, 17(153):1–43, 2016.

[31] Ilya Sutskever, James Martens, George Dahl, and Geoffrey Hinton. On the importance of initialization and momentum in deep learning. In *International conference on machine learning*, pages 1139–1147, 2013.

[32] N.G. van Kampen. The Diffusion Approximation for Markov Processes. In I. Lamprecht and A. I. Zotin, editors, *Thermodynamics and Kinetics of Biological Processes*, pages 181–195. Walter de Gruyter & Co., 2013.

# A On weak approximation of diffusion process to stochastic gradient descent

In the main text, we have considered the stochastic gradient descent (SGD) iteration

$$x^{(k)} = x^{(k-1)} - \eta \nabla f(x^{(k-1)}, \zeta_k) , \ x^{(0)} = x_0 \in \mathbb{R}^d , \tag{A.1}$$

in which $\zeta_k \sim \mathcal{D}$ is an i.i.d. sequence of random variables with the same distribution $\mathcal{D}$, and $\eta > 0$ is the learning rate. Usually $\eta$ is taken to be quite small. We approximate (A.1) by the diffusion process

$$d\boldsymbol{X}(s) = -\nabla F(\boldsymbol{X}(s))ds + \sqrt{\eta}\mathbf{S}(\boldsymbol{X}(s))\,d\boldsymbol{B}_s , \ \boldsymbol{X}(0) = \mathbf{x}_0 \in \mathbb{R}^d . \tag{A.2}$$

Such an approximation can justified as in the proof of Theorem 3.2. However, the problem with this approximation is in that (a) the approximation works on $[0, T]$ for fixed $T > 0$; (b) The approximation error is characterized by a constant $C > 0$ that may not be dimension–free. (c). To gain $O(\eta)$ weak approximation, $\mathbf{S}$ can be arbitrary smooth bounded functions.

In this appendix we aim at explaining that the limit process (A.2) can be viewed as a weak approximation to the discrete iteration (A.1) and the term $\sqrt{\eta}\mathbf{S}(\boldsymbol{X}(s))\,d\boldsymbol{B}_s$ is essential to capture the $O(\sqrt{\eta})$ fluctuation.

We propose the following way of understanding the approximation, that is essentially adapted from classical monographs [2, 3, 26]. Let us introduce a deterministic process that can be characterized by an ordinary differential equation

$$d\boldsymbol{Y}(s) = -\nabla F(\boldsymbol{Y}(s))ds , \ \boldsymbol{Y}(0) = x_0 \in \mathbb{R}^d . \tag{A.3}$$

It turns out, that as $\eta > 0$ tends to zero, we have strong norm convergence

$$\lim_{\eta \to 0} \max_{0 \le k \le T/\eta} \|x^{(k)} - \boldsymbol{Y}(\eta k)\|_{\mathbb{R}^d} = 0 . \tag{A.4}$$

Indeed, on larger time scales, the above approximation can be realized in a weak topology specified below. For each fixed $T > 0$, let $C_{[0,T]}(\mathbb{R}^d)$ be the space of continuous functions from the interval $[0, T]$ to $\mathbb{R}^d$. For any function $\phi \in C_{[0,T]}(\mathbb{R}^d)$, we equip the space $C_{[0,T]}(\mathbb{R}^d)$ with norm $\|\phi\|_{C_{[0,T]}(\mathbb{R}^d)} = \sup_{0 \le t \le T} |\phi(t)|$. Let $C_{[0,\infty)}(\mathbb{R}^d)$ be the space of continuous functions from $[0, \infty)$ to $\mathbb{R}^d$. For any function $\phi \in C_{[0,\infty)}(\mathbb{R}^d)$, we equip the space $C_{[0,\infty)}(\mathbb{R}^d)$ with norm $\|\phi\|_{C_{[0,\infty)}(\mathbb{R}^d)} = \sum_{n=1}^{\infty} \frac{1}{2^n} \|\phi\|_{C_{[0,n]}(\mathbb{R}^d)}$. Under the norm $\|\cdot\|_{C_{[0,\infty)}(\mathbb{R}^d)}$, the space $C_{[0,\infty)}(\mathbb{R}^d)$ becomes a separable Banach space.

For a family of functions $\phi_n \in C_{[0,\infty)}(\mathbb{R}^d)$ and the function $\phi \in C_{[0,\infty)}(\mathbb{R}^d)$, we say that $\phi_n \rightharpoonup \phi$ weakly in $C_{[0,\infty)}(\mathbb{R}^d)$ if and only if for every linear functional $F : C_{[0,\infty)}(\mathbb{R}^d) \to \mathbb{R}$ that is continuous in the sense that $F(\phi_n - \phi) \to 0$ if $\|\phi_n - \phi\|_{C_{[0,\infty)}(\mathbb{R}^d)} \to 0$ as $n \to \infty$, we have $F(\phi_n - \phi) \to 0$ for every such linear functional $F$ as $n \to \infty$.

Let us modify the trajectory $x^{(k)}$ in (A.1) into a continuous function $x^\eta(t)$ by linearly interpolate between points $((k-1)\eta, x^{(k-1)})$ and $(k\eta, x^{(k)})$, $k = 1, 2, ...$, i.e.,

$$(t, x^\eta(t)) = \left(t, \left(\frac{t}{\eta} - k + 1\right) x^{(k)} + \left(k - \frac{t}{\eta}\right) x^{(k-1)}\right)$$

for $(k-1)\eta \leq t < k\eta$. In this way the function $x(t)$ can be viewed as an element in $C_{[0,\infty)}(\mathbb{R}^d)$. Moreover, the solutions $\boldsymbol{X}(t)$ in (A.2) and $\boldsymbol{Y}(t)$ in (A.3) are also elements in $C_{[0,\infty)}(\mathbb{R}^d)$. We claim the following two results.

**Theorem 4.** *For any sequence $\eta_n \to 0$, we have $x^{\eta_n}(t) \rightharpoonup \boldsymbol{Y}(t)$ as $n \to \infty$ in the space $C_{[0,\infty)}(\mathbb{R}^d)$.*

The understanding of the diffusion approximation $\boldsymbol{X}(t)$ to $x(t)$ can be achieved via normal deviations. To see that, one can consider the rescaled process

$$\zeta^\eta(t) = \frac{x^\eta(t) - \boldsymbol{Y}(t)}{\sqrt{\eta}} \ . \tag{A.5}$$

**Theorem 5.** *For any sequence $\eta_n \to 0$, we have $\zeta^{\eta_n}(t) \rightharpoonup \boldsymbol{Z}(t)$ as $n \to \infty$ in the space $C_{[0,\infty)}(\mathbb{R}^d)$. The process $\boldsymbol{Z}(t)$ is characterized by the stochastic differential equation*

$$d\boldsymbol{Z}(s) = M(s)\boldsymbol{Z}(s)ds + \boldsymbol{S}(\boldsymbol{Y}(s))d\widetilde{W}_s \ , \ \boldsymbol{Z}(0) = 0 \in \mathbb{R}^d \ . \tag{A.6}$$

*in which $\widetilde{W}_s$ is a standard Brownian motion in $\mathbb{R}^d$, and $M(s)$ is a bounded $d \times d$ matrix function.*

The above theorems can be proved via standard methods in ODE approximation (see [3]) and normal deviation theory (see [2], Part II, Chapter 4, Theorem 7). We omit the details here.

By (A.6), we can say that we have an approximate expansion (such expansions appear in physics literatures and are called Van–Kampen's approximation [32]) of the form

$$x(t) \approx \boldsymbol{Y}(t) + \sqrt{\eta} \int_0^t \boldsymbol{S}(\boldsymbol{Y}(s))d\widetilde{W}_s \ . \tag{A.7}$$

Notice that by (A.3), we have

$$\boldsymbol{Y}(t) = x_0 - \int_0^t \nabla F(\boldsymbol{Y}(s))ds \ .$$

Therefore we have an expansion

$$x(t) \approx x_0 - \int_0^t \nabla F(\boldsymbol{Y}(s))ds + \sqrt{\eta} \int_0^t \boldsymbol{S}(\boldsymbol{Y}(s))d\widetilde{W}_s \ .$$

Note that by (A.5) and (A.6) the processes $x(t)$ and $\boldsymbol{Y}(t)$ are close at order $O(\sqrt{\eta})$. From here, we have approximately

$$x(t) \approx x_0 - \int_0^t \nabla F(x(s))ds + \sqrt{\eta} \int_0^t \boldsymbol{S}(x(s))d\widetilde{W}_s \ .$$

This justifies the approximation of $\boldsymbol{X}(t)$ in (A.2) to $x(t)$.

# B  Detailed discussion regarding Theorem 3

Since $D$ is bounded, we can modify the values of $\mathbf{S}$ and $F$ outside $D$ so that they and their derivatives are bounded in the whole space, which clearly does not change the hitting time.

Recall that $\mathbf{x} = 0$ is a nondegenerate saddle point or a nondegenerate local maximum point. Consider the dynamics system given by

$$\frac{d}{ds}\mathbf{x}(s) = -\nabla F(\mathbf{x}(s)), \ \mathbf{x}(0) = \mathbf{x}$$

Let

$$\mathcal{W}^S = \{\mathbf{x} \in D : \lim_{s \to \infty} \mathbf{x}(s) = 0, \mathbf{x}(0) = \mathbf{x} \neq 0\}, \tag{B.1}$$

which is called the stable manifold of the dynamics system and define

$$\theta(\mathbf{x}) = \inf\{s > 0 : \mathbf{x}(s) \in \partial D\}. \tag{B.2}$$

In [24, Theorem 2.2], the following asymptotics for the mean exit time were proved, which applies to local maximum point as well:

**Proposition 3.** *If* $\mathbf{x} \in \mathcal{W}^S \cup \{0\}$*, then*

$$\lim_{\eta \to 0} \frac{1}{\log \eta^{-1}} \mathbb{E}_{\mathbf{x}} \mathcal{T}^\eta = \frac{1}{2\gamma_1}.$$

*If* $\mathbf{x} \in D \setminus (\mathcal{W}^S \cup \{0\})$*, then*

$$\lim_{\eta \to 0} \mathbb{E}_{\mathbf{x}} \mathcal{T}^\eta = \theta(\mathbf{x}).$$

**Remark 2.** *The escaping time* (4.12) *from the unstable critical point can be understood intuitively as following: in the ball* $B(0, \eta^{0.5})$*, the Brownian motion dominates and the time that the process arrives at the boundary is* $s = \mathcal{O}(1)$*. From* $\partial B(0, \eta^{0.5})$ *to* $\partial B(0, 1)$*, the convection term dominates and it is essentially* $\mathbf{X}' = \gamma_1 \mathbf{X}$ *and hence time spent for the second stage is* $s \sim \log(\eta^{-1})$*.*

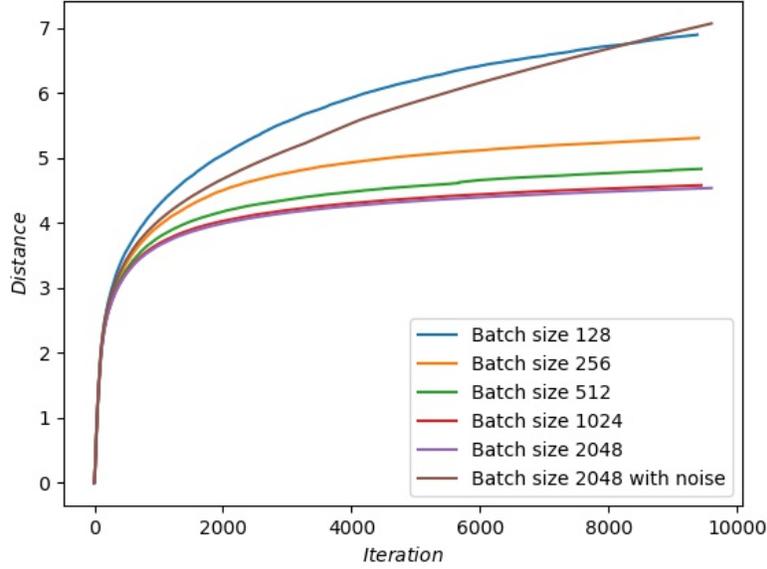

Figure 1: Euclidean distance of weight vector from the initial weight vector for each iteration. We choose 128, 256, 512, 1024, 2048 as batch size and set learning rate $\eta = 0.1$. For the conjecture we set $\sigma = 0.00054$ to make the curve similar with small batch method's curve.

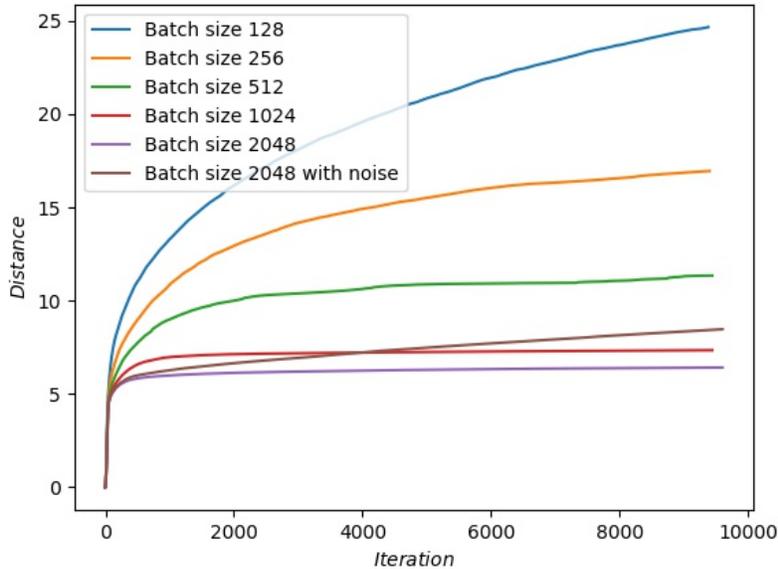

Figure 2: Euclidean distance of weight vector from the initial weight vector for each iteration. We choose 128, 256, 512, 1024, 2048 as batch size and set learning rate $\eta = 0.1$, momentum $\rho = 0.9$. For the conjecture we set $\sigma = 0.00054$ to increase randomness level of large batch method.

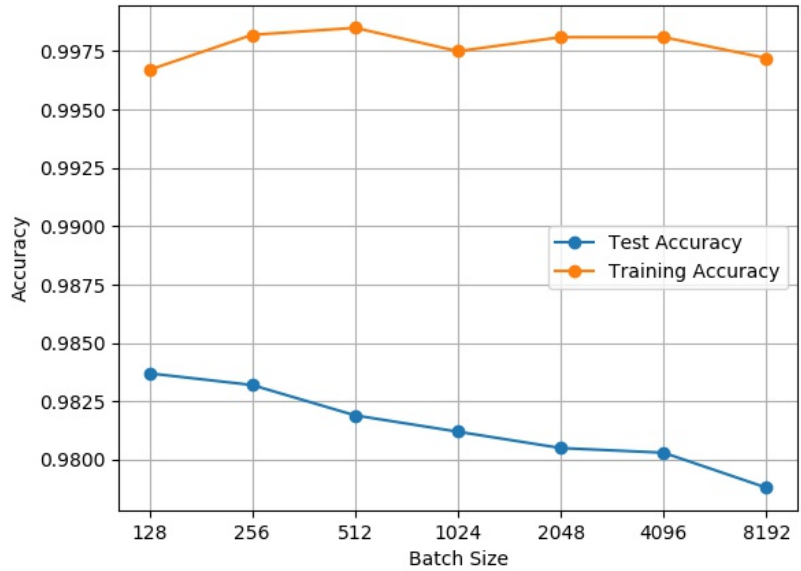

Figure 3: The accuracy of the test and training set evaluated as the number of batch size. We set $\eta = 0.1$, momentum $\rho = 0$ and the same number of iterarions on MNIST

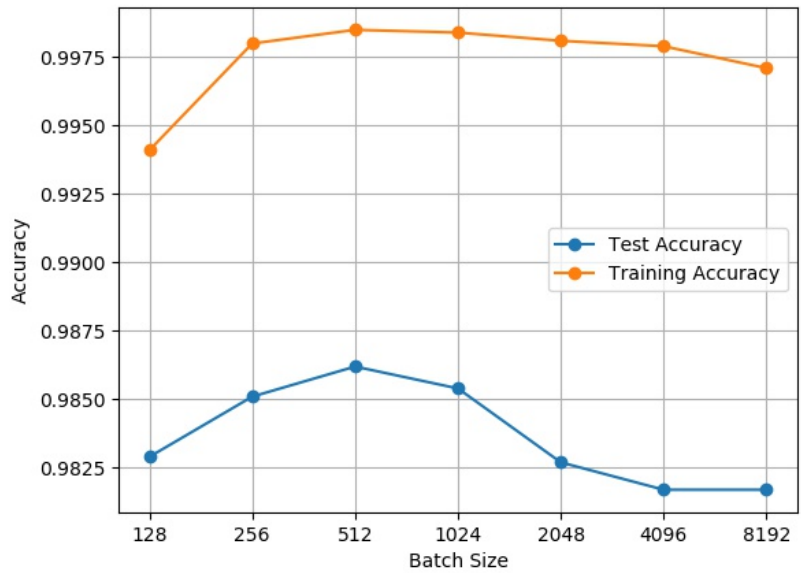

Figure 4: The accuracy of the test and training set evaluated as the number of batch size. We set $\eta = 0.1$, momentum $\rho = 0.9$ and same number of iterations on MNIST

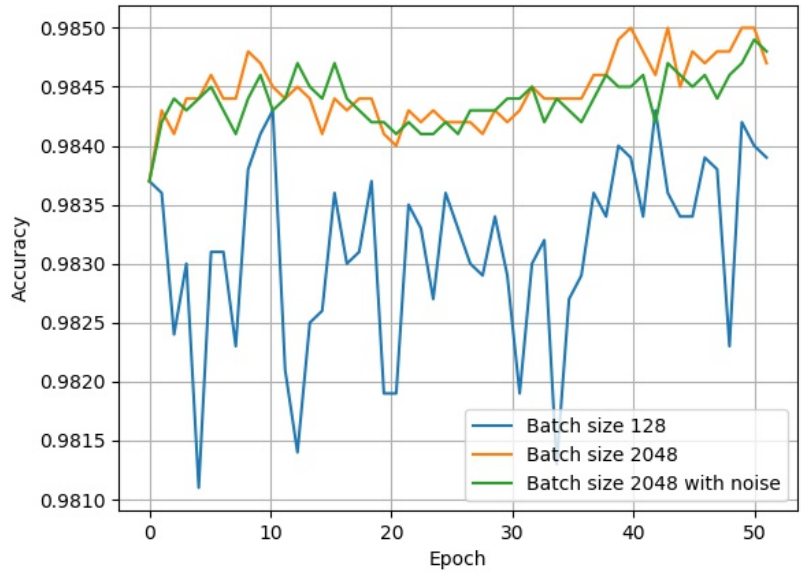

Figure 5: The accuracy of the test set evaluated as the number of epoch. The best result of small batch method as the initial weight and set $\eta = 0.1$

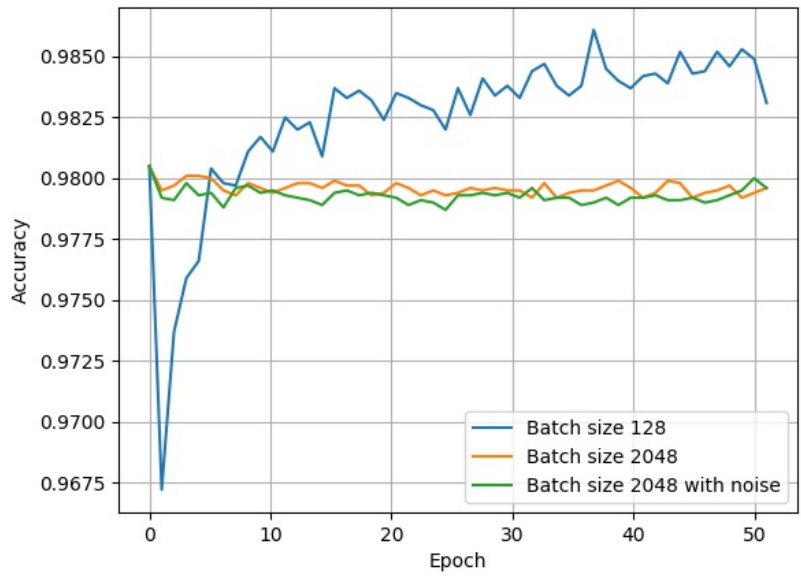

Figure 6: The accuracy of the test set evaluated as the number of epoch. The best result of large batch method as the initial weight and set $\eta = 0.1$